\documentclass[letterpaper, 10 pt, conference]{ieeeconf}
\usepackage{times}

\IEEEoverridecommandlockouts                              

\usepackage{multicol}
\usepackage[bookmarks=true]{hyperref}
\usepackage{xcolor}
\usepackage{subfigure}
\usepackage{graphicx}
\usepackage{amsmath}
\usepackage{wrapfig}
\usepackage{flushend}
\DeclareMathAlphabet\mathbfcal{OMS}{cmsy}{b}{n}

\usepackage{algorithm}
\usepackage{algpseudocode}

\usepackage{amssymb}
\usepackage{pifont}
\newcommand{\cmark}{\ding{51}}%
\newcommand{\xmark}{\ding{55}}%

\title{\LARGE \bf
ARROCH: Augmented Reality for Robots Collaborating with a Human 
}

\author{Kishan Chandan, Vidisha Kudalkar, Xiang Li, Shiqi Zhang
\thanks{Chandan, Kudalkar, and Zhang are with SUNY Binghamton.}
\thanks{Email: {\tt\footnotesize \{kchanda2, vkudalk1, zhangs\}@binghamton.edu}}
\thanks{Li is with OPPO US Research Center.} 
\thanks{Email: {\tt\footnotesize xiang.li@oppo.com}}
}

\begin{document}

\maketitle
\thispagestyle{empty}
\pagestyle{empty}

\begin{abstract}

Human-robot collaboration frequently requires extensive communication, e.g., using natural language and gesture. 
Augmented reality (AR) has provided an alternative way of bridging the communication gap between robots and people.
However, most current AR-based human-robot communication methods are unidirectional, focusing on how the human adapts to robot behaviors, and are limited to single-robot domains.
In this paper, we develop \emph{AR for Robots Collaborating with a Human} (ARROCH), a novel algorithm and system that supports bidirectional, multi-turn, human-multi-robot communication in indoor multi-room environments. 
The human can see through obstacles to observe the robots' current states and intentions, and provide feedback, while the robots' behaviors are then adjusted toward human-multi-robot teamwork. 
Experiments have been conducted with real robots and human participants using collaborative delivery tasks. 
Results show that ARROCH outperformed a standard non-AR approach in both user experience and teamwork efficiency. 
In addition, we have developed a novel simulation environment using Unity (for AR and human simulation) and Gazebo (for robot simulation). 
Results in simulation demonstrate ARROCH's superiority over AR-based baselines in human-robot collaboration. 

\end{abstract}

\IEEEpeerreviewmaketitle

\section{Introduction}

Robots are increasingly present in everyday environments, such as warehouses, hotels, and airports, but human-robot collaboration~(\textbf{HRC}) is still a challenging problem. 
As a consequence, for instance, the work zones for robots and people in warehouses are separated~\cite{wurman2008coordinating}; and hotel delivery robots barely communicate with people until the moment of delivery~\cite{ivanov2017adoption}. 
When people and robots work in shared environments, it is vital that they communicate to collaborate with each other to avoid conflicts, leverage complementary capabilities, and facilitate the smooth accomplishment of tasks. 
In this paper, we aim to enable people and robot teams to efficiently and accurately communicate their current states and intentions toward collaborative behaviors. 

Augmented reality (\textbf{AR}) technologies focus on visually overlaying information in an augmented layer over the real environment to make the objects interactive~\cite{azuma2001recent}. 
AR has been applied to human-robot systems, where people can visualize the state of the robot in a visually enhanced form~\cite{green2007augmented}. 
Most existing research on AR-based human-robot collaboration focuses on the visualization of robot status, and how people leverage the augmented information to adapt to robot behaviors, resulting in unidirectional communication~\cite{walker2018communicating, chadalavada2015s}. 
Another observation is that many of those methods from the literature were limited to human-single-robot, in-proximity scenarios~\cite{ganesan2018better,luebbers2019augmented}. 
In this paper, we focus on \textbf{human-multi-robot}, \textbf{beyond-proximity} settings, where the robots need to collaborate with each other and with people at the same time, in indoor multi-room environments. 
Our developed algorithm (and system) is called \textbf{ARROCH}, short for \emph{AR for robots collaborating with a human}. 

ARROCH supports \textbf{bidirectional} human-robot communication. 
On the one hand, ARROCH enables the human to visualize the robots' current states, e.g., their current locations, as well as their intentions (planned actions), e.g., to enter a room. 
For instance, a human might see through an opaque door via AR to ``X-ray'' a mobile robot waiting outside along with its planned motion trajectory. 
On the other hand, leveraging a novel AR interface, ARROCH allows the human to give feedback to the robots' planned actions, say to temporarily forbid the robots from entering an area. 
Accordingly, the robots can incorporate such human feedback for replanning, avoiding conflicts, and constructing synergies at runtime. 
ARROCH supports beyond-proximity communication by visually augmenting the robots' states and intentions, which is particularly useful in indoor multi-room environments. 

ARROCH (algorithm and system), as the \textbf{first contribution} of this paper, has been evaluated in simulation and using real robots. 
The robots help people move objects, and the human helps the robots open doors (the robots do not have an arm, and cannot open the doors).
Results from the real-world experiment suggest that ARROCH significantly improves the efficiency of human-robot collaboration, compared with a standard non-AR baseline. 
The simulation platform is constructed using Unity~\cite{juliani2020unity} for simulating the AR interface and human behaviors, and using Gazebo~\cite{koenig2004design} for simulating robot behaviors. 
To the best of our knowledge, this is the first open-source simulation platform for simulating AR-based human-multi-robot behaviors, which is the \textbf{second contribution} of this research.
In the simulation, ARROCH significantly improved human-robot teamwork efficiency in comparison to baseline methods with different communication mechanisms. 

    
    

\begin{figure*}
\begin{center}
    \vspace{.8em}
    \includegraphics[width=15.5cm]{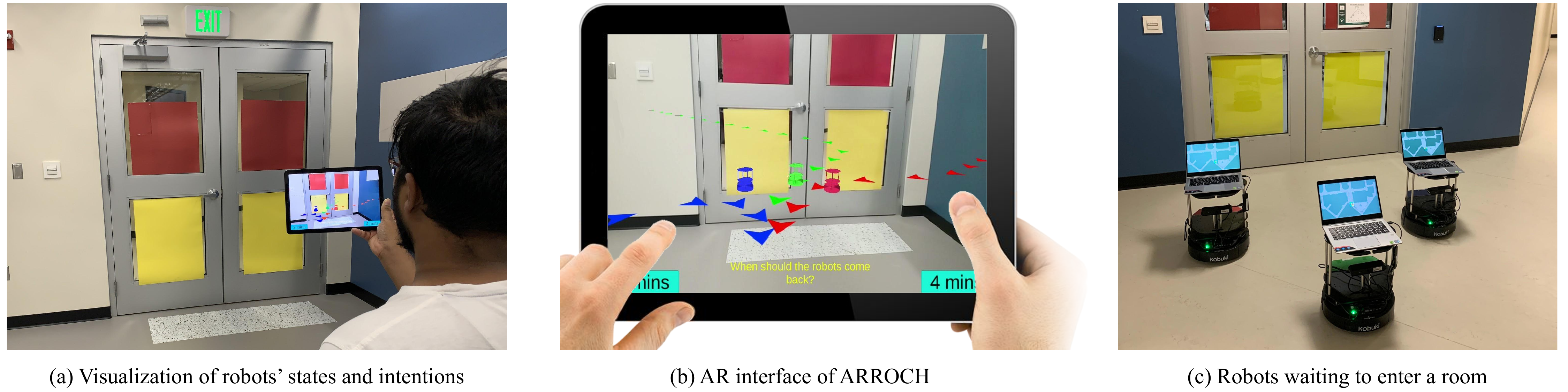}
    \vspace{-.8em}
    \caption{The AR interface of ARROCH enables the human to visualize the robots' current states (their current locations in this example) using 3D robot avatars, and their intentions (entering the room) using trajectory markers. 
    ARROCH also enables the human to give feedback to robot plans. 
    In this example, the human can use the interactive buttons in the bottom corners to indicate he could not help open the door in a particular time frame, say ``4 minutes.'' }
    \label{fig:MainOverviewARROCH}
\end{center}
\vspace{-1.8em}
\end{figure*}

    
    

\section{Related Work}
\label{sec:related}

\noindent
\textbf{Human-Robot Communication Modalities: }
Humans and robots prefer different communication modalities. 
While humans employ natural language and gestures, the information in digital forms, such as text-based commands, is more friendly to robots. 
Researchers have developed algorithms to bridge the human-robot communication gap using natural language~\cite{chai2014collaborative,thomason2015learning,matuszek2013learning,amiri2019augmenting}
and vision~\cite{Waldherr2000,nickel2007visual,yang2007gesture}. 
Despite those successes, augmented reality (AR) has its unique advantages of providing a communication medium with potentially less ambiguity and higher bandwidth~\cite{azuma1997survey}.
AR helps in elevating coordination through communicating spatial information, e.g., through which door a robot is coming into a room and how (i.e., the planned trajectory), when people and robots share a physical environment. 
Researchers are increasingly paying attention to AR-based human-robot systems~\cite{williams2019virtual}. 
We use an AR interface for human-robot communication, where the human can directly visualize and interact with the robots' current and planned actions. 

\vspace{.5em}
\noindent
\textbf{Projection-based Communication: }
One way of delivering spatial information related to the local environment, referred to as ``projection-based AR,'' is through projecting the robot's state and motion intention to the humans using visual cues~\cite{chadalavada2015s,Park:2009:RPA:1514095.1514146,4373930}.
For instance, researchers used an LED projector attached to the robot to show its planned motion trajectories, allowing the human partner to respond to the robot's plan to avoid possible collisions~\cite{7354195}.
While projection-based AR systems facilitate human-robot communication about spatial information, they have the requirement that the human must be in close proximity to the robot.
We develop our AR-based framework that inherits the benefits of spatial information from the projection-based systems while alleviating the proximity requirement, and enabling bidirectional communication.



\vspace{.5em}
\noindent
\textbf{Unidirectional AR-based: }
More recently, researchers have developed frameworks to help human operators visualize the motion-level intentions of unmanned aerial vehicles~(UAVs) using AR~\cite{walker2018communicating}, and visualize a robot arm's planned actions in the car assembly tasks~\cite{ganesan2018better}.
One common limitation of those systems is their unidirectional communication, i.e., their methods only convey the robot's intentions to the human but do not support the communication the other way around. 
In comparison, ARROCH supports bidirectional communication, and has been applied to human-multi-robot, multi-room collaboration domains.

\vspace{.5em}
\noindent
\textbf{Bidirectional AR-based: }
Early research on AR-based human-robot interaction~(HRI) has enabled a human operator to interactively plan, and optimize robot trajectories~\cite{583833, green2010evaluating}.
The AR-based systems have also been used to improve the teleoperation of collocated robots~\cite{hedayati2018improving}.
Recently, researchers have studied how AR-based visualizations can help the shared control of robotic systems~\cite{brooks2020visualization}.
Most relevant to this paper is a system that supports a human user to visualize the robot's sensory information, and planned trajectories, while allowing the robot to ask questions through an AR interface~\cite{muhammad2019creating}.
In comparison to their work on single-robot domains, ARROCH supports human-multi-robot collaboration, where both multi-robot and human-robot teamwork is supported. 
More importantly, our robots are equipped with the task (re)planning capability, which enables the robots to respond to the human feedback by adjusting their task completion strategy leading to collaborative behaviors within human multi-robot teams.
The key properties of a sample of existing methods and ARROCH are summarized in Table~\ref{tab:related_work_table}.

\begin{table}[t] \footnotesize
    \vspace{.5em}
    \caption{A summary of the key properties of a sample of existing AR-based human-robot communication methods. 
    }
    \vspace{-.5em}
    \centering
    \begin{tabular}{|l|c|c|c|c|c|} 
    \hline
    & Beyond & Bidir. & Multiple & Task \\ 
    & proximity & comm.  & robots  & planning\\
    \hline
         \cite{chadalavada2015s, 7354195, ganesan2018better} & \xmark & \xmark & \xmark & \xmark \\
    \hline
         \cite{Park:2009:RPA:1514095.1514146, 4373930, brooks2020visualization,muhammad2019creating} & \xmark & \cmark & \xmark & \xmark\\
    
     \hline
     \cite{hedayati2018improving, green2010evaluating} & \cmark & \cmark & \xmark & \xmark\\
     \hline
     ARROCH (ours) & \cmark & \cmark & \cmark & \cmark\\
     \hline
    \end{tabular}
    \label{tab:related_work_table}
    \vspace{-1.5em}
\end{table}

\section{ARROCH: Algorithm and System}

In this section, we describe ARROCH (\emph{AR for robots collaborating with a human}), an algorithm and system that utilizes AR technologies to enable bidirectional, multi-turn, beyond-proximity communication to enable collaborative behaviors within human-multi-robot teams. 
Fig.~\ref{fig:MainOverviewARROCH} shows our novel AR interface, which enables the human to visualize the robots' current states, and their intentions (planned actions), while at the same time supporting the human to give feedback to the robots. 
The robots can then use the feedback to adjust their plans as necessary. 
Next, we describe the ARROCH algorithm, and then its system implementation.



\subsection{The ARROCH Algorithm}
\label{sec:alg}

The input of Algorithm~\ref{alg:arn_algorithm} includes, $\boldsymbol{\mathcal{S}}$, a set of initial states of $N$ robots, where $s_{i} \in \boldsymbol{\mathcal{S}}$ is the initial state of the $i^{th}$ robot; $\boldsymbol{\mathcal{G}}$, a set of goal states, where $g_{i} \in \boldsymbol{\mathcal{G}}$ is a goal state of the $i^{th}$ robot.\footnote{Strictly speaking, tasks are defined as goal conditions~\cite{ghallab2004automated}. Here we directly take goal states as input for simplification. } 
A multi-robot task planner, $\mathcal{P}^{t}$, and a motion planner, $\mathcal{P}^{m}$ are also provided as input.

\hfill\break
\vspace{-2.95em}
\begin{algorithm}[h] \footnotesize
\caption{ARROCH}\label{alg:arn_algorithm}
\textbf{Input}: $\boldsymbol{\mathcal{S}}$, $\boldsymbol{\mathcal{G}}$, $\mathcal{P}^{t}, \mathcal{P}^{m}$ 
\begin{algorithmic}[1]

\State {Initialize empty lists: $\boldsymbol{\omega} \xleftarrow{} \emptyset$; $\boldsymbol{\mathcal{I}} \xleftarrow{} \emptyset$; $\boldsymbol{\mathcal{C}} \xleftarrow{} \emptyset$}


\State{$\boldsymbol{P} = \mathcal{P}^{t}(\boldsymbol{\mathcal{S}}, \boldsymbol{\mathcal{G}}, \boldsymbol{\mathcal{C}})$, where $p \in \boldsymbol{P}$ is a task plan (an action sequence) for one robot, and $|\boldsymbol{P}|=N$} \label{line:generate_initial_plans}\algorithmiccomment{Task planner $\mathcal{P}^{t}$ in Section \ref{sec:planner}}



\While{$\sum_{p\in \boldsymbol{P}}|p| > 0$} \label{line:main_while_loop}

\For{$i \in [0, 1, \cdots, N$-$1]$}\label{line:for_loop_robots}
\If{$p_{i}$ is \textbf{not} empty} \label{line:goal_reached}
\State $a_{i} \xleftarrow{} p_{i}$.front()\algorithmiccomment{Obtain current robot action} \label{line:see_front_action}
\State Obtain $i^{th}$ robot's current configuration: $\boldsymbol{\omega}_{i} \xleftarrow{}\theta(i)$\label{line:get_pose_robots}
\State Update the $i^{th}$ robot's current state $s_{i}$ using $\boldsymbol{\omega}_{i}$
\State $\boldsymbol{\mathcal{I}}_{i} \xleftarrow{} \mathcal{P}^{m}(\boldsymbol{\omega}_{i}, a_{i})$  \label{line:generate_trajectory} 
\State The $i^{th}$ robot follows $\boldsymbol{\mathcal{I}}_{i}$ using a controller

\EndIf

\EndFor\label{line:for_loop_robots_end}

\State{$\lambda \xleftarrow{} \mathcal{V}(\boldsymbol{\omega}, \boldsymbol{\mathcal{I}})$} \label{line:Visualization_Agent} \algorithmiccomment{Visualizer $\mathcal{V}$ defined in Section \ref{sec:visualizer}}

\State{Collect feedback, $H$, from human, where the human gives feedback based on task completion status and $\lambda$}\label{line:getHumanFeedback}

\State{$\boldsymbol{\mathcal{C}} \xleftarrow{} \mathcal{R}(H)$} \label{line:get_constraints}\algorithmiccomment{Restrictor $\mathcal{R}$ defined in Section \ref{sec:restrictor}}

\State{$\boldsymbol{P} \xleftarrow{} \mathcal{P}^{t}(\boldsymbol{\mathcal{S}}, \boldsymbol{\mathcal{G}}, \boldsymbol{\mathcal{C}})$}\label{line:generateNewPlans} \algorithmiccomment{Update plans}



\EndWhile \label{line:end_while}

\end{algorithmic}
\end{algorithm}

ARROCH starts by initializing: an empty list of robot configurations (in \emph{configuration space}, or C-Space), $\boldsymbol{\omega}$, where each configuration is in the simple form of $\langle x,y,\theta \rangle$ for our mobile robots, an empty list, $\boldsymbol{\mathcal{I}}$, to store the intended trajectories of the $N$ robots in C-Space ($N=|\boldsymbol{\mathcal{S}}|$), and an empty list to store the activated constraints, $\boldsymbol{\mathcal{C}}$.

In Line~\ref{line:generate_initial_plans}, ARROCH generates a set of $N$ plans, $\boldsymbol{P}$, using multi-robot task \textbf{Planner} $\mathcal{P}^{t}$.
Entering the while-loop (Lines~\ref{line:main_while_loop}-\ref{line:end_while}), ARROCH runs until $\sum_{p\in \boldsymbol{P}}|p| > 0$ is false, meaning that all robots have an empty plan, i.e., all tasks have been completed.
In each iteration, ARROCH enters a for-loop for updating the current configuration and the intended trajectories of the robots yet to reach their goal states (Line~\ref{line:for_loop_robots}-\ref{line:for_loop_robots_end}).
The ${\theta}$ function in Line~\ref{line:get_pose_robots} returns the configuration of the $i^{th}$ robot, which is stored at $\boldsymbol{\omega}_{i}$.
In Line~\ref{line:generate_trajectory}, $\mathcal{P}^{m}$ generates the intended trajectory of the $i^{th}$ robot, $\boldsymbol{\mathcal{I}}_{i}$, to implement action $a_i$. 
After the for-loop, the set of robot configurations and robots' intended trajectories, are passed to \textbf{Visualizer} (Section~\ref{sec:visualizer}), represented by the $\mathcal{V}$ function, which renders a single frame, $\lambda$, on the AR interface (Line~\ref{line:Visualization_Agent}).

The AR interface of ARROCH allows the human to give feedback on (the AR-based visualization of) the robots' plans.
ARROCH obtains the human feedback, $H$ (Line~\ref{line:getHumanFeedback}), and then passes it on to \textbf{Restrictor}~(Line~\ref{line:get_constraints}). 
Restrictor generates a set of activated constraints in logical form, $\boldsymbol{\mathcal{C}}$, that can be processed by Planner in Line~\ref{line:generateNewPlans} (details in Sections~\ref{sec:restrictor} and~\ref{sec:planner}).

\vspace{.5em}
\noindent
\textbf{Remarks: }
Leveraging a novel AR interface, ARROCH enables bidirectional, multi-turn, beyond-proximity communication toward collaborative behaviors within human-multi-robot teams. 
ARROCH is particularly useful in domains with poor visibility, e.g., multi-room indoor environments. 
The multi-robot task replanning capability enables the robots to collaborate with each other, while responding to human feedback by adjusting their task completion strategies.

\begin{figure}[t]
\begin{center}
\vspace{0.5em}
    \includegraphics[width=.49\textwidth]{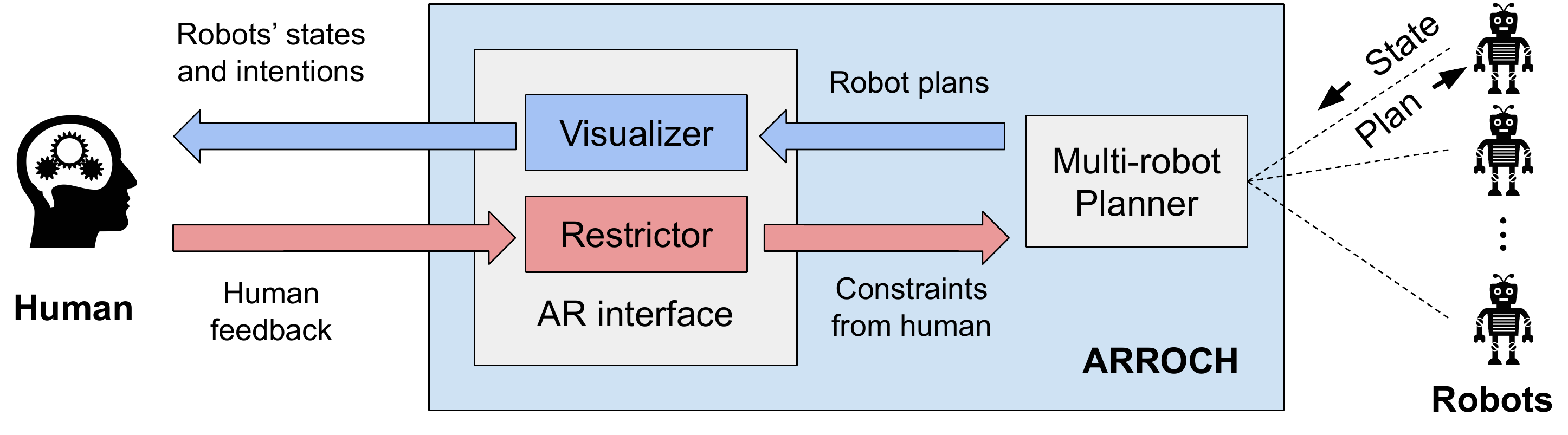}
    \vspace{-1.5em}
    \caption{Key components of our ARROCH algorithm and system: \emph{Visualizer} and \emph{Restrictor} for visualizing robot's intention (for people) and collecting human feedback (for robots) respectively, and \emph{Planner} for computing one action sequence for each robot. }
    \label{fig:framework}
\end{center}
\vspace{-1.5em}
\end{figure}

Fig.~\ref{fig:framework} shows an overview of ARROCH. 
In the following subsections, we describe how \textbf{Restrictor} converts human feedback into symbolic constraints, how \textbf{Planner} incorporates the constraints to plan for the robots, and how the robots' current and planned actions are visualized in \textbf{Visualizer} -- all within our ARROCH system. 

\subsection{Restrictor for Constraint Generation}
\label{sec:restrictor}

ARROCH realizes human-multi-robot collaboration by adding human-specified symbolic constraints into the multi-robot task planner. 
Within the task planning context, a \textbf{Constraint} is in form of modal-logic expressions, which should be true for the state-trajectory produced during the execution of a plan~\cite{gerevini2009deterministic}. 
However, naive people have difficulties in directly encoding constraints in logical forms. 
ARROCH leverages the \emph{graphical user interface} (GUI) of AR to take human feedback, with which Restrictor generates logical constraints. 
We use Answer Set Programming (ASP) for encoding constraints~\cite{lifschitz2008answer,zhang2015mobile}. 
A constraint in ASP is technically a STRIPS-style ``headless'' rule: 

    \vspace{.5em}
    \texttt{:- B.}
    \vspace{.5em}
    
\noindent
where \texttt{B} is a conjunction of literals. It conveys the intuition of a constraint: satisfying \texttt{B} results in a contradiction. 
There is rich literature on ASP-based logic programming that provides more technical details~\cite{baral2003knowledge,gelfond2014knowledge}.

We predefine a constraint library $F$, which is presented to people through the AR-based GUI. 
The human can select constraints $H \subseteq F$ to give feedback on the robots' plans (Line~\ref{line:getHumanFeedback} in Algorithm~\ref{alg:arn_algorithm}). 
The form of human feedback can be flexible and task-dependent. 
For instance, in tasks that involve navigation, the human can forbid the robots from entering area \texttt{A} in step \texttt{I} using the following constraint, 

    \vspace{.5em}
    \texttt{:- in(A,I), area(A), step(I).}
    \vspace{.5em}
    
\noindent
and in collaborative assembly tasks, the human can reserve a tool for a period of time. 
ARROCH uses Restrictor $\mathcal{R}$ to convert $H$ into a set of constraints, $\boldsymbol{\mathcal{C}}$, which is the logical form of human feedback. 
Next, we describe how $\boldsymbol{\mathcal{C}}$ (from human) is incorporated into multi-robot task planning. 

\subsection{Multi-Robot Task Planner}
\label{sec:planner}

In this subsection, we focus on the multi-robot task planner, $\mathcal{P}^{t}$ (referred to as \textbf{Planner}), that computes one task plan for each robot from the team of robots, while also considering the human feedback (Lines~\ref{line:generate_initial_plans} and~\ref{line:generateNewPlans} in Algorithm~\ref{alg:arn_algorithm}). 
The input of $\mathcal{P}^{t}$ include, a set of robot initial states, $\boldsymbol{\mathcal{S}}$; a set of robot goal states, $\boldsymbol{\mathcal{G}}$; and a set of constrained resources, $\boldsymbol{\mathcal{C}}$.



Jointly planning for multiple agents to compute the optimal solution is NP-hard~\cite{sharon2015conflict}. 
We adapt an \emph{iterative inter-dependent planning} (IIDP) approach to compute joint plans for multiple robots~\cite{jiang2019multi}. 
IIDP starts with independently computing one plan for each robot toward completing their non-transferable tasks. 
After that, in each iteration, IIDP computes an optimal plan for a robot under the condition of other robots' current plans (i.e., this planning process depends on existing plans). 
While IIDP does not guarantee global optimality, it produces desirable trade-offs between plan quality and computational efficiency.

In the implementation of ARROCH, we use ASP to encode the action knowledge (for each robot) that includes the description of five actions: \texttt{approach}, \texttt{opendoor}, \texttt{gothrough}, \texttt{load}, and \texttt{unload}.
For instance, 

    \vspace{.5em}
    \texttt{open(D,I+1) :- opendoor(D,I), }
    
    \texttt{~~~~~~~~~~~~~~~door(D), step(I).}
    \vspace{.5em}
    
\noindent    
states that executing action \texttt{opendoor(D,I)} causes door \texttt{D} to be open at the next step. 
The following states that a robot cannot execute the \texttt{opendoor(D)} action, if it is not facing door \texttt{D} at step \texttt{I}. 
Such constraints are provided by Restrictor, as described in Section~\ref{sec:restrictor}. 

\vspace{.5em}
\texttt{:- opendoor(D,I), not facing(D,I).}
\vspace{.5em}

The output of $\mathcal{P}^{t}$ is $\boldsymbol{P}$, a set of task plans (one for each robot). 
For instance, $p_i \in \boldsymbol{P}$ can be in the form of: 

    \vspace{.5em}
    \texttt{load(O,0). approach(D,1). opendoor(D,2).\!\!\!}
    
    \texttt{gothrough(D,3). unload(O,4).}
    \vspace{.5em}
    
\noindent   
suggesting the $i^{th}$ robot to pick up object \texttt{O}, approach door \texttt{D}, enter a room through \texttt{D}, and drop off the object. 
Next, we describe how ARROCH handles the visualization of robots' current states and their planned actions through an AR interface.

\subsection{AR Visualizer} 
\label{sec:visualizer}

ARROCH uses an AR interface to bridge the communication gap in human-multi-robot teams, where \textbf{Visualizer}~$\mathcal{V}$ in Line~\ref{line:Visualization_Agent} of Algorithm~\ref{alg:arn_algorithm} plays a key role.
The input of $\mathcal{V}$ includes the robots' current configurations in C-Space, $\boldsymbol{\omega}$, and their intended motion trajectories, $\boldsymbol{\mathcal{I}}$. 
The output of $\mathcal{V}$ are spatially visualizable 3D objects which are augmented over the real world.
To accurately overlay virtual objects over the real world, AR devices need to localize itself in 3D environments. 
ARROCH assumes the availability of a set of landmark objects with known 3D poses (position and orientation). 
Using visual localization, ARROCH computes the pose of the AR device, and then accordingly augments visual information over the mobile device. We use a tablet for AR, though ARROCH is compatible with other mobile devices, e.g., head-mounted displays. 
Fig.~\ref{fig:MainOverviewARROCH}~(b) presents an example of the illustrative visualization. 


In this section, we describe the ARROCH algorithm and system, the first contribution of this paper.
Next, we delineate a novel simulation platform (our \textbf{second contribution} of this paper) for prototyping and evaluating AR-based human-multi-robot collaboration algorithms. 

\begin{figure}[t]
\begin{center}
    \subfigure[Gazebo: office]
    {\includegraphics[height=2.4cm]{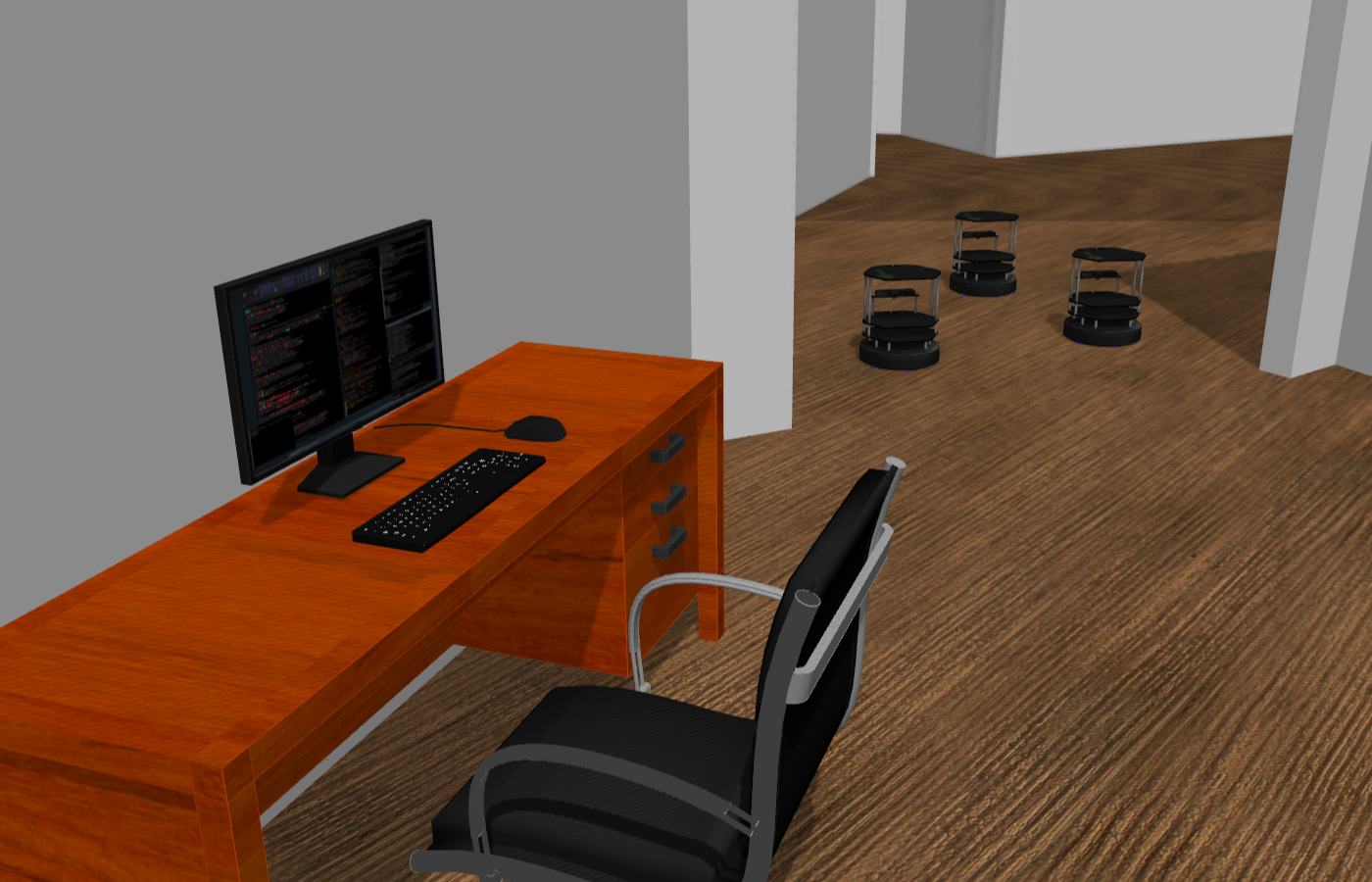}
    \label{fig:gazebo_robots_new}}
    \subfigure[Gazebo: robots]
    {\includegraphics[height=2.4cm]{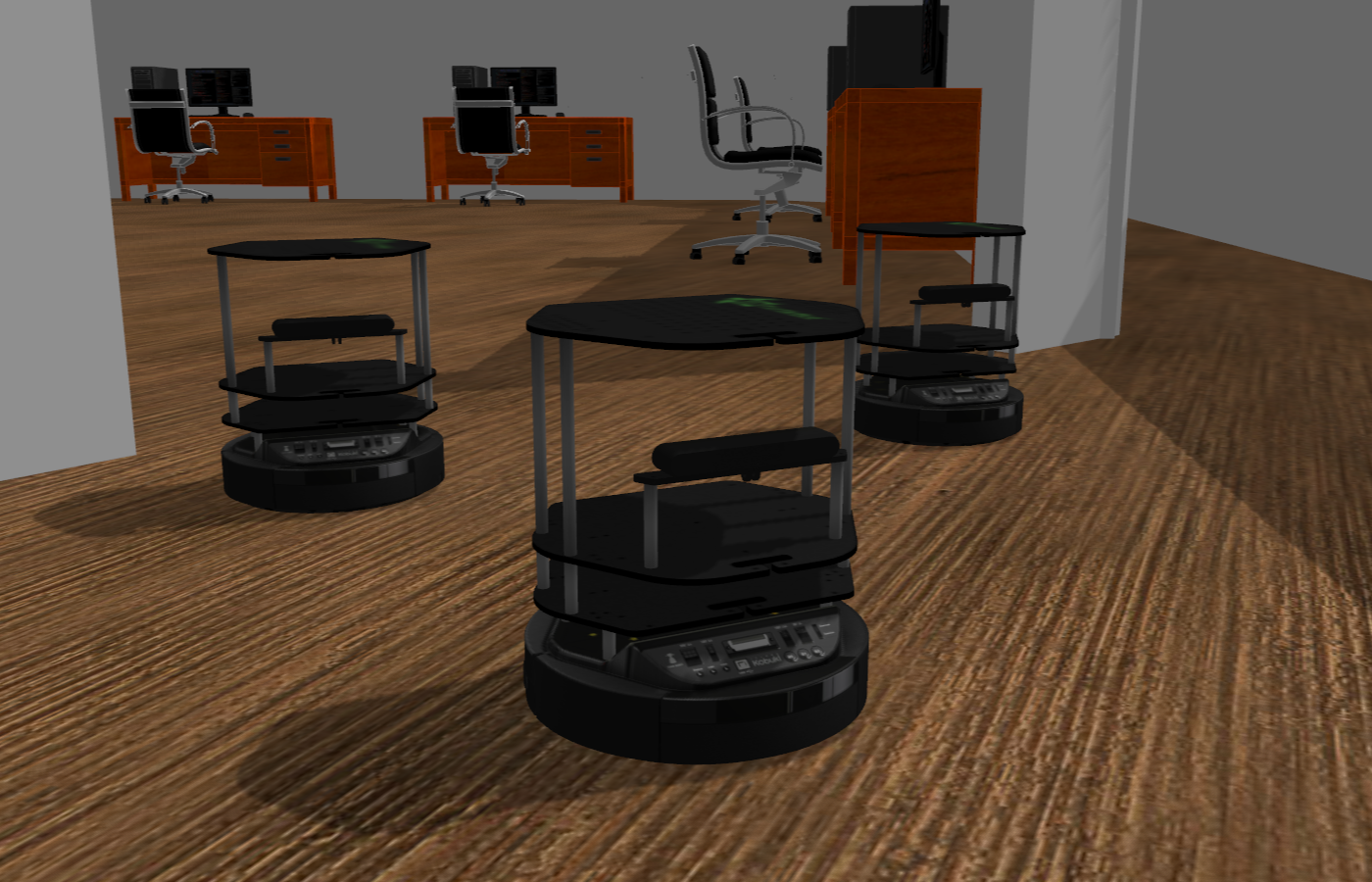}
    \label{fig:zoomed_robots_new}}
    \subfigure[Unity: office]
    {\includegraphics[height=2.4cm]{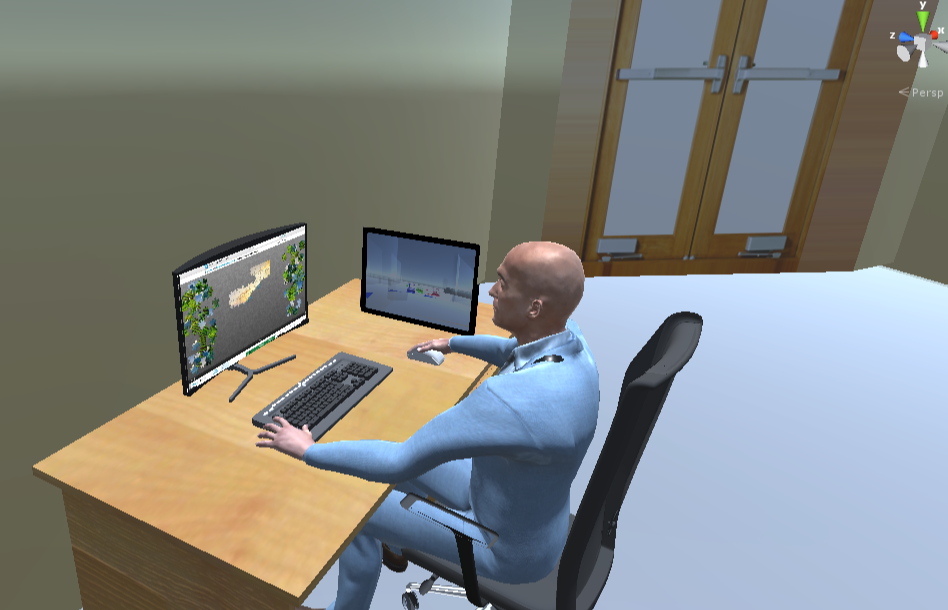}
    \label{fig:unity_table_chair}}
    \subfigure[Unity: robots]
    {\includegraphics[height=2.4cm]{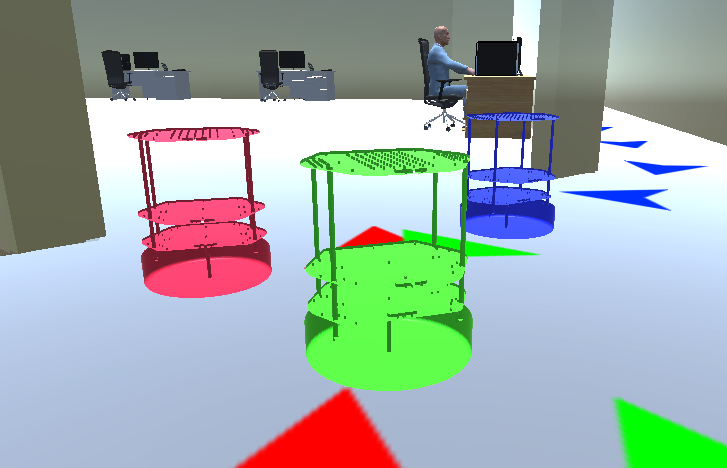}
    \label{fig:unity_zoomed_robots}}
    \subfigure[Unity: AR (1st person POV)]
    {\includegraphics[height=2.4cm]{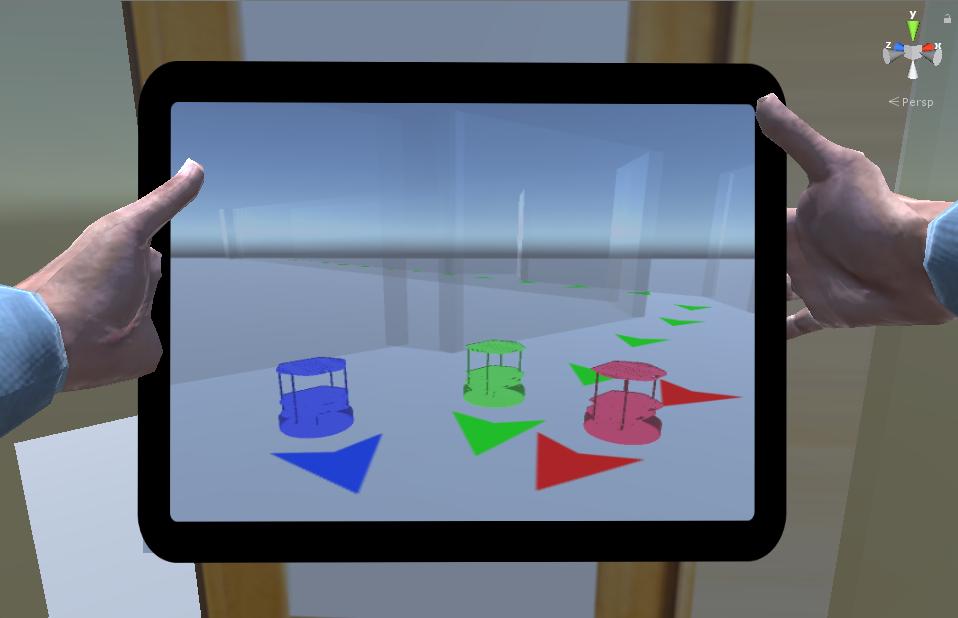}
    \label{fig:unity_zoomed_ar_interface}}    
    \subfigure[Unity: AR (3rd person POV)]
    {\includegraphics[height=2.4cm]{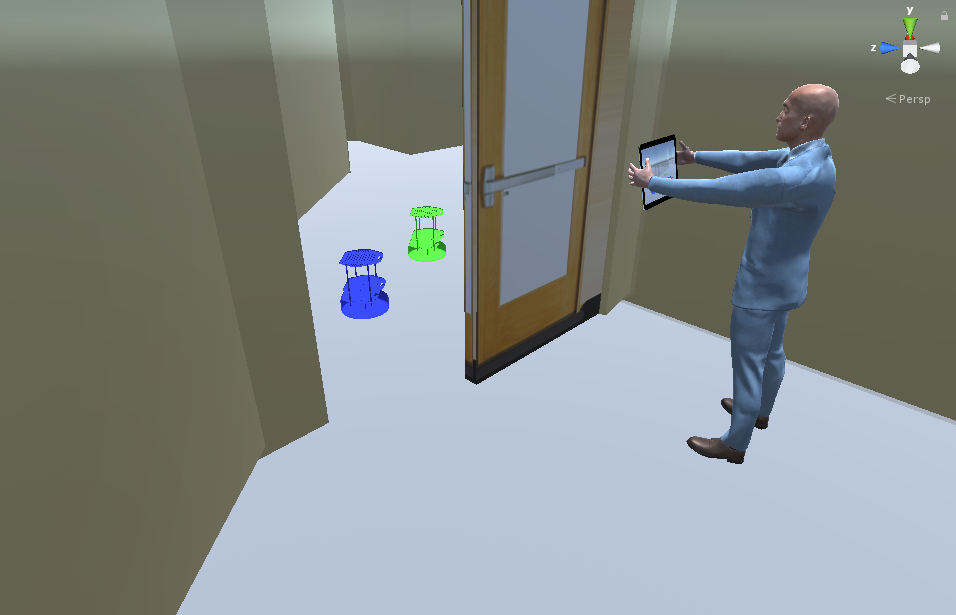}
    \label{fig:unity_human_holding_ar_interface}}
    
    \caption{(a)-(b),~Gazebo environment for simulating multi-robot behaviors; 
    (c)-(d),~Unity environment for simulating human behaviors and her AR-based interactions with the robots; and 
    (e)-(f),~Simulated AR interface. 
    }
    \label{fig:sim_figs_new}
    \end{center}
    \vspace{-1.5em}
\end{figure}

\section{SUGAR2: \textbf{S}imulation with \textbf{U}nity and \textbf{G}azebo for \textbf{A}ugmented \textbf{R}eality and \textbf{R}obots}
\label{sec:sim}

Our simulation platform has been \textbf{open-sourced} under the name of \textbf{SUGAR2}.\footnote{https://github.com/kchanda2/SUGAR2}
SUGAR2 is built on Gazebo~\cite{koenig2004design} for simulating robot behaviors, and Unity~\cite{juliani2020unity} for simulating human behaviors and AR-based interactions.
Fig.~\ref{fig:sim_figs_new} shows the Gazebo and Unity environments, including a simulated AR interface.
Gazebo does not support the simulation of AR-based interactions, and Unity is weak in simulating robot behaviors, which motivated the development of SUGAR2. 

\section{Experiments}
\label{sec:exp}

We have conducted experiments in simulation and using real robots.
We aim to evaluate \textbf{two hypotheses}: I)~ARROCH improves the overall efficiency in human-multi-robot team task completion, determined by the slowest agent of a team, in comparison to AR-based methods that do not support bidirectional human-robot communication; and II)~ARROCH produces better experience for non-professional users in human-multi-robot collaboration tasks in comparison to non-AR methods. 
\textbf{Hypothesis-I} was evaluated in simulation and using real robots, whereas \textbf{Hypothesis-II} was only evaluated based on questionnaires from real human-robot teams. 

In both simulation and real-world environments, a human-multi-robot team works on \textbf{collaborative delivery tasks}. 
The robots help people move objects from different places into a storage room, while the human helps the robots open a door that the robots cannot open by themselves. 
At the same time, the human has her own dummy task (solving Jigsaw puzzles). 
Tasks are non-transferable within the human-multi-robot team. 
The tasks and experiment settings are shared by simulation and real-world experiments.

\begin{figure}[t]
\begin{center}
    \vspace{0.4em}
        \includegraphics[width=.48\textwidth]{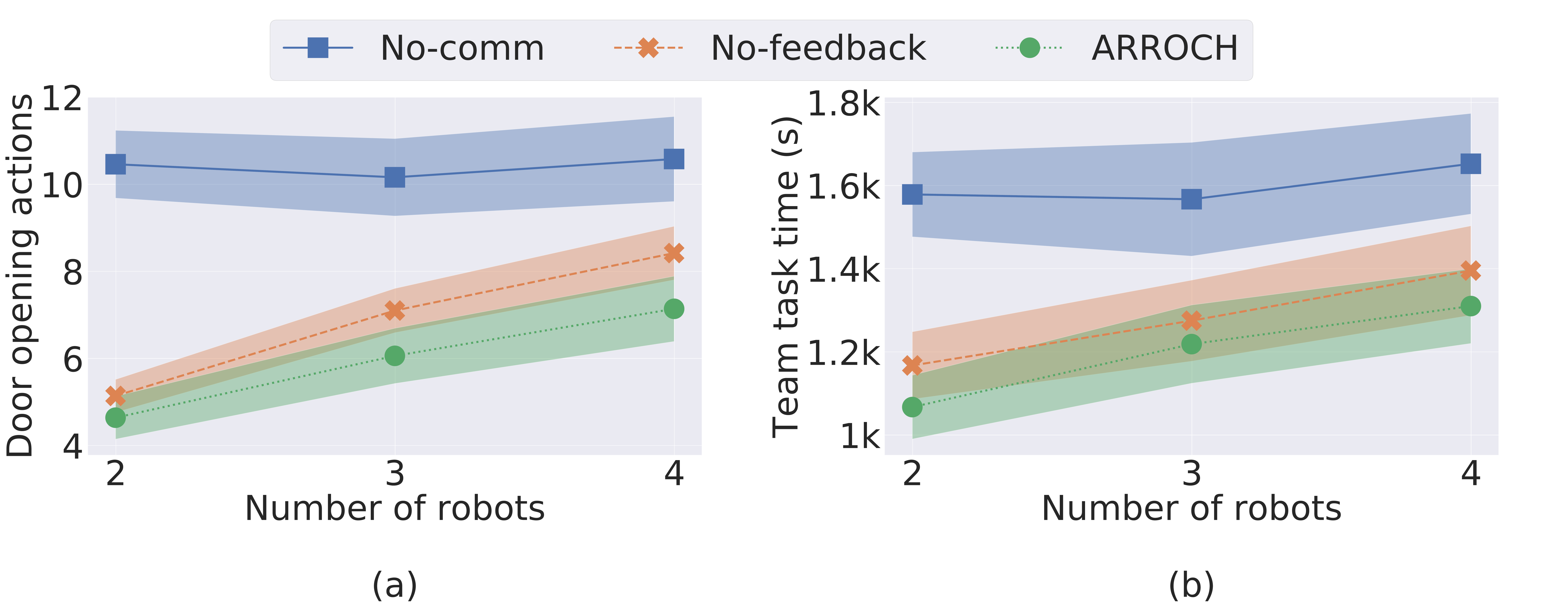}
    \vspace{-2em}
    \caption{With different numbers of robots, ARROCH enables the human-robot team to: (a)~reduce the human's workload, measured by the number of door-opening actions, and (b)~improve the task-completion efficiency, at the same time.
    The total delivery workload (the number of objects to be delivered) is proportionally increased when the number of robots increases. }
    \label{fig:sim_exp2}
\end{center}
\vspace{-2em}
\end{figure}

\subsection{Simulation Experiments using SUGAR2}
\label{sec:exp_sim}

We use the SUGAR2 environment (Section~\ref{sec:sim}) to simulate human and robot behaviors, and their AR-based interactions. 
ARROCH has been compared with \textbf{two baselines}:
\begin{itemize}
    \item \textbf{No-feedback: } It is the same as ARROCH, except that Lines~\ref{line:getHumanFeedback}-\ref{line:get_constraints} in Algorithm~\ref{alg:arn_algorithm} are deactivated. 
    As a result, the human can see the robots' current states and planned actions, but cannot provide feedback to the robots. 
    \item \textbf{No-comm: } It is the same as No-feedback, except that Line~\ref{line:Visualization_Agent} is deactivated. 
    As a result, AR-based communication is completely disabled. 
\end{itemize}



\vspace{.5em}
\noindent
\textbf{Three types of human behaviors} are simulated in SUGAR2. 
$a^H_0$,~\emph{Work on her own task (simplified as sitting still in Unity)}; 
$a^H_1$,~\emph{Help robots open the door}; and 
$a^H_2(N)$,~\emph{Indicate unavailability in the next N minutes}. 

When the AR-based visualization is activated (ARROCH and No-feedback), the human's door-opening behavior is independently triggered by each waiting robot, meaning that the human is more likely to open the door when more robots are waiting for the human to open the door. 
For No-comm, in every minute, there is a probability (0.6 in our case) that the human goes to open the door. 
To realistically simulate human behaviors, we added noise (Gaussian) into the human's task completion time. 
Before the human completing her task, we randomly sample time intervals during which the human will be completely focusing on her own task. 
At the beginning of such intervals, there is a 0.9 probability that the human takes action $a^H_2(N)$, i.e., indicating this ``no interruption'' time length to the robots in ARROCH.

\vspace{.5em}
\noindent{\bf{Different Number of Robots:}}
Fig.~\ref{fig:sim_exp2} presents the results in simulation experiments, where we varied the number of robots in the human-multi-robot team. 
Each robot's workload is fixed. 
In our case, each robot needs to deliver three objects. 
Fig.~\ref{fig:sim_exp2}~(a) shows that ARROCH produced the lowest number of door-opening actions which directly reduces the human's workload.
Fig.~\ref{fig:sim_exp2}~(b) shows that ARROCH produced the best performance in the human-multi-robot team's task-completion time, which is determined by the slowest agent.
The results support that Hypothesis-I is valid in teams of different sizes.

\begin{figure}[t]
\vspace{0.4em}
\begin{center}
        \includegraphics[width=.48\textwidth]{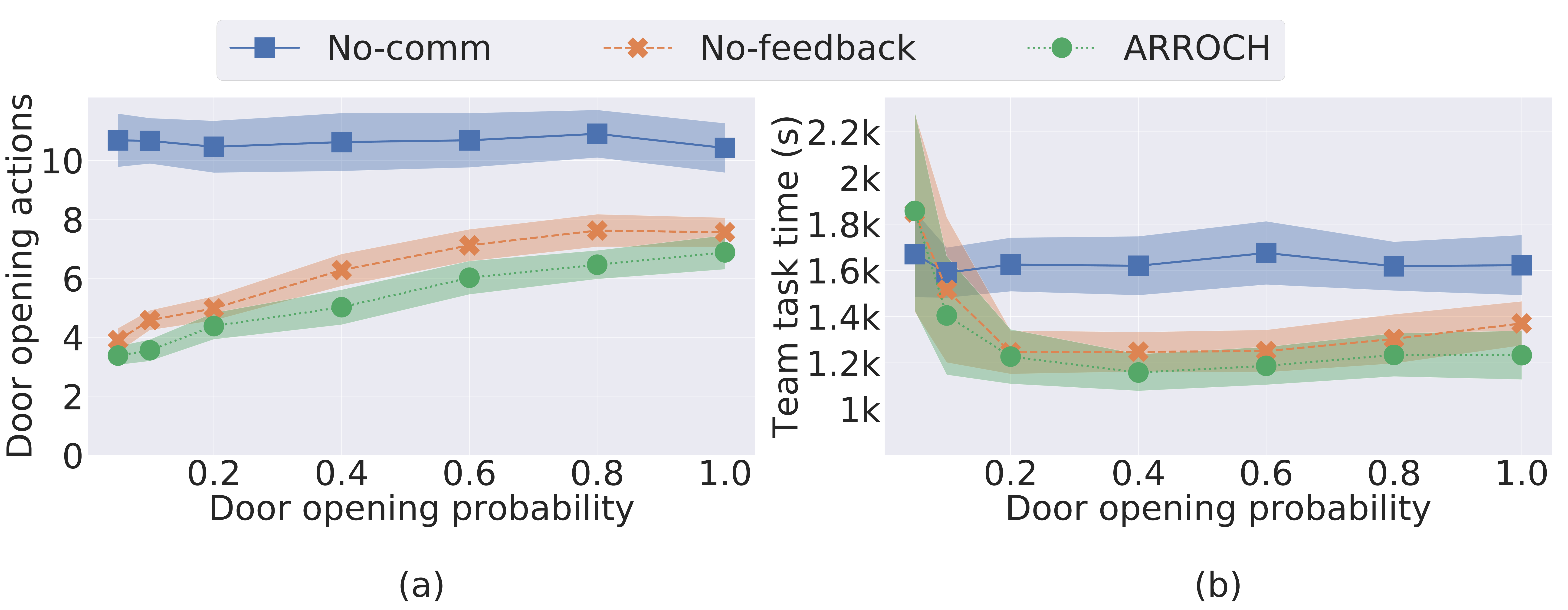}
    \vspace{-2em}
    \caption{With different door-opening probabilities (the lower it is, the lazier the human is), ARROCH enables the team to: (a)~reduce the number of door-opening actions, and (b)~improve the task-completion efficiency, as long as the human is reasonably cooperative (i.e., willing to help the robots in $\geq 0.1$ probability). }
    \label{fig:sim_exp1}
\end{center}
\vspace{-1.5em}
\end{figure}

\vspace{.5em}
\noindent{\bf{Human Laziness:}}
Fig.~\ref{fig:sim_exp1} shows the results of an experiment, where we evaluated ARROCH's performance under different human behaviors. 
We varied the human's laziness level that is inversely proportional to the probability that the human opens the door after she sees (through AR) a robot waiting outside. 
From the results, we see that ARROCH significantly reduced the number of door-opening actions (Left subfigure), while producing the best performance in task completion efficiency (Right subfigure), except for situations where the ``door opening probability'' is very low, meaning that the human is very lazy. 
This observation makes sense, because ARROCH only \emph{encourages} lazy people to help robots (so does No-feedback) because the door opening behavior is triggered based on the number of visualized robots.
On the other hand, ``No-comm'' forces people to open the door after every minute.
The results support that Hypothesis-I is valid as long as the human is reasonably cooperative. 

\begin{figure*}[t]
\vspace{-0.25em}
\begin{center}
    \subfigure[Individual completion time]
    {\includegraphics[height=3.2cm]{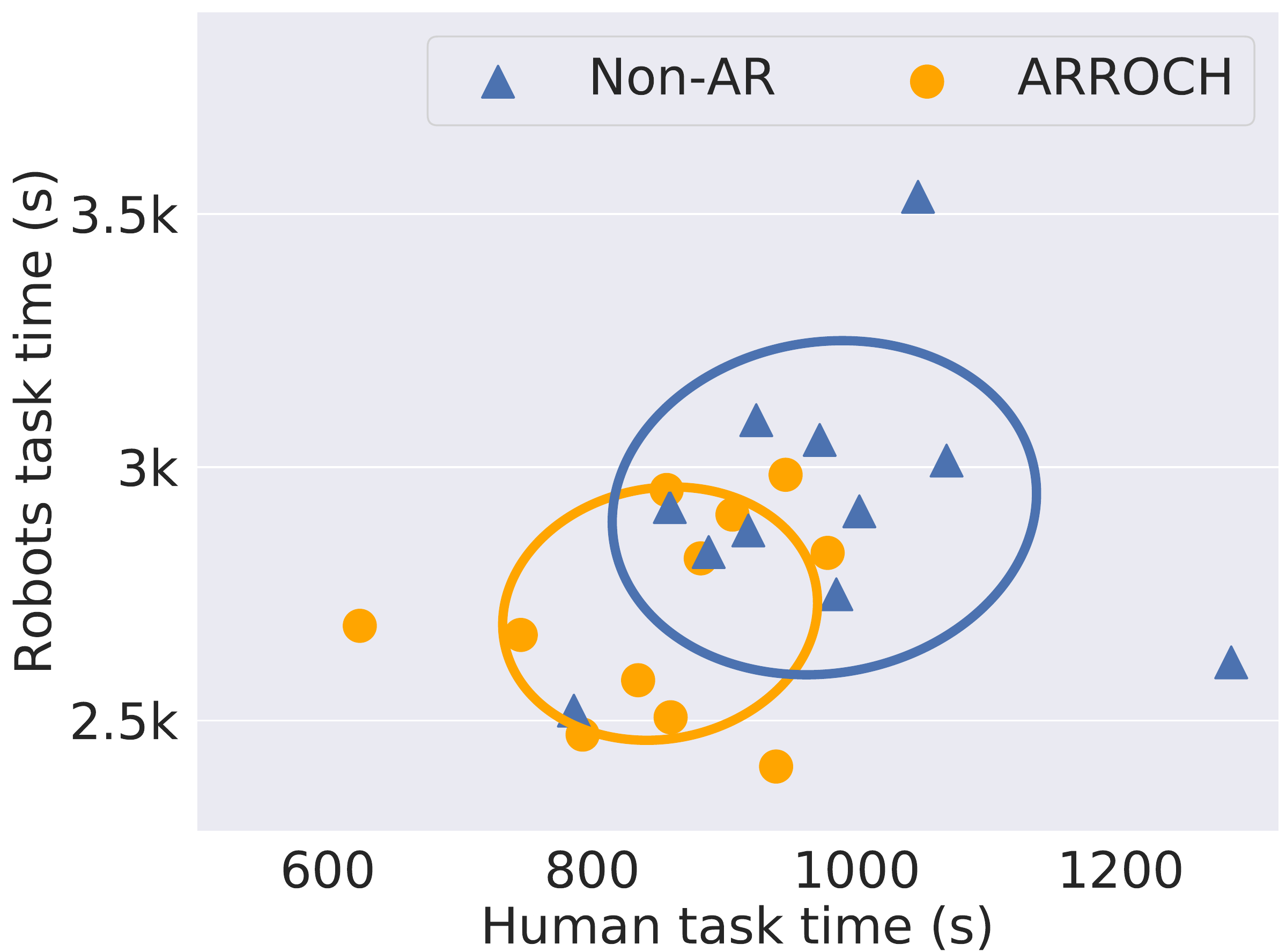}
    \label{fig:scatter_plot}}
    \hspace{0.5cm}
    \subfigure[Team completion time (histogram)]
    {\includegraphics[height=3.2cm]{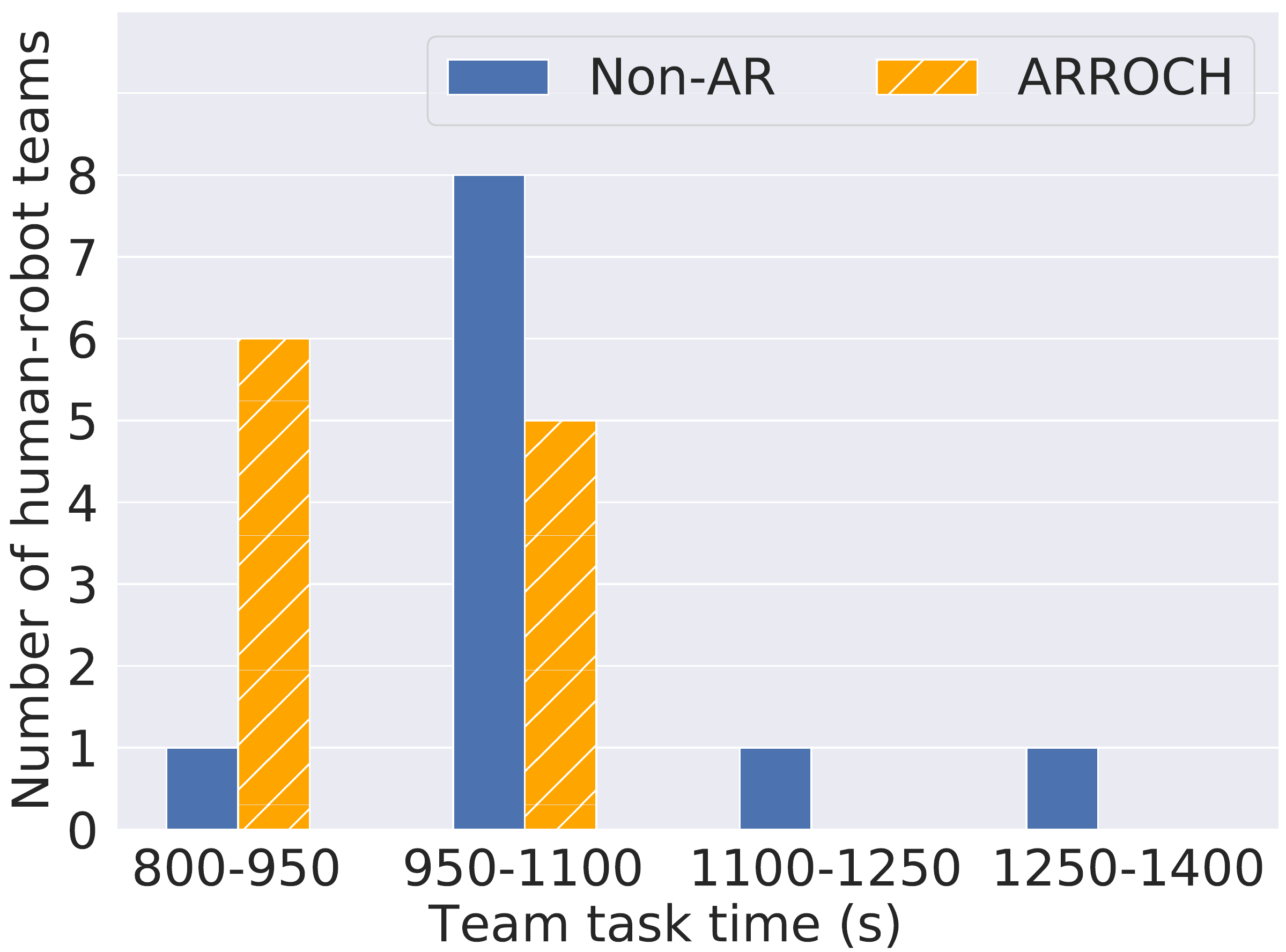}
    \label{fig:histogram_real}}
    \hspace{0.5cm}
    \subfigure[Questionnaire scores]
    {\includegraphics[height=3.2cm]{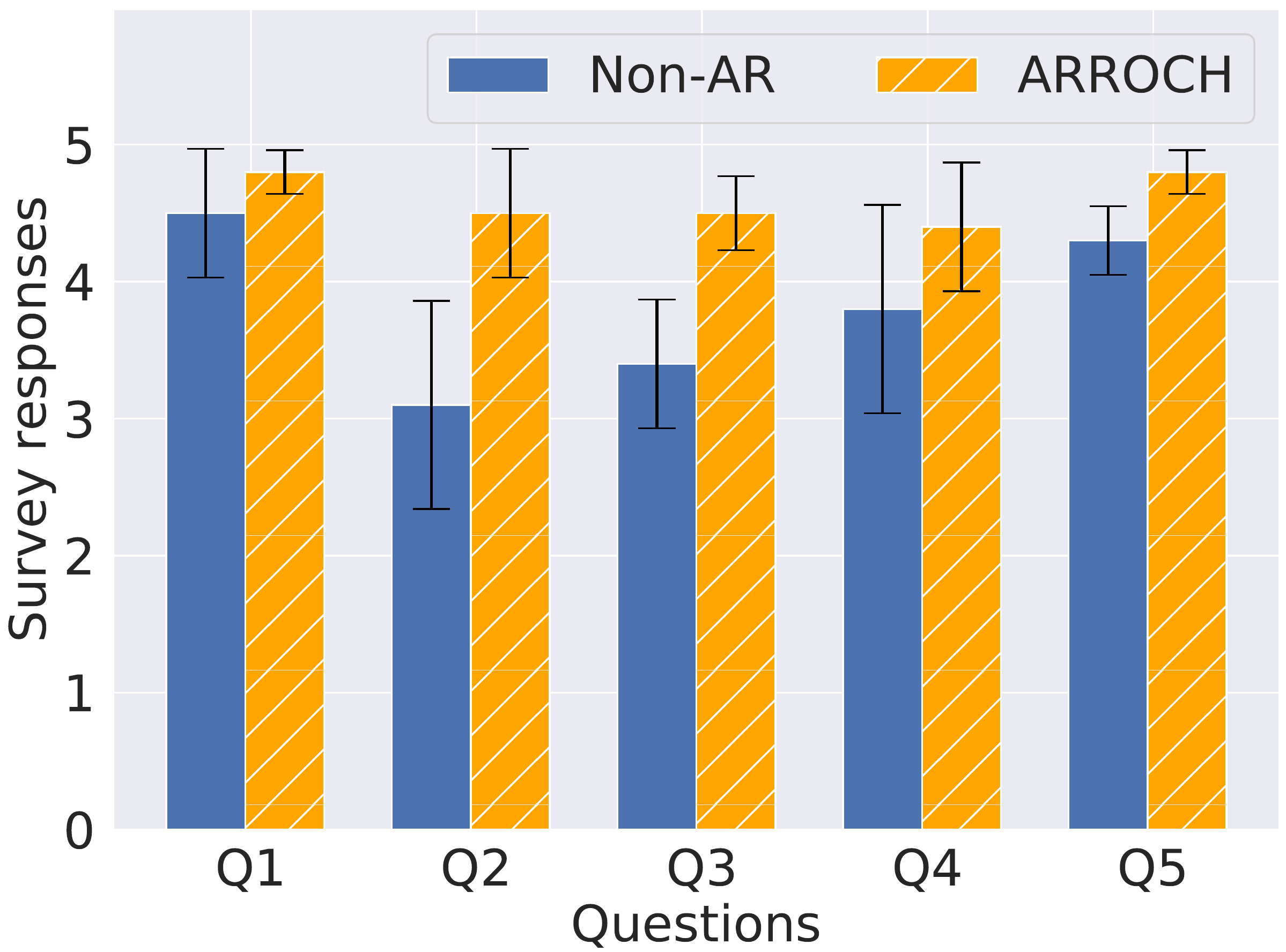}
    \label{fig:bar_graph}}
    \vspace{-.8em}
    \caption{(a)~ARROCH performed better than a traditional non-AR baseline (Rviz-based) in individual task completion time;
    (b)~ARROCH performed better, c.f., non-AR, in team task completion time; and
    (c)~ARROCH produces a better user experience based on results collected from user questionnaires.
    }
    \label{fig:results}
    \end{center}
    \vspace{-1.5em}
\end{figure*}

\subsection{Real-world Experiments}

\begin{figure}[t]
    \begin{center}
    \subfigure[Human participant]
    {\includegraphics[height=2.7cm]{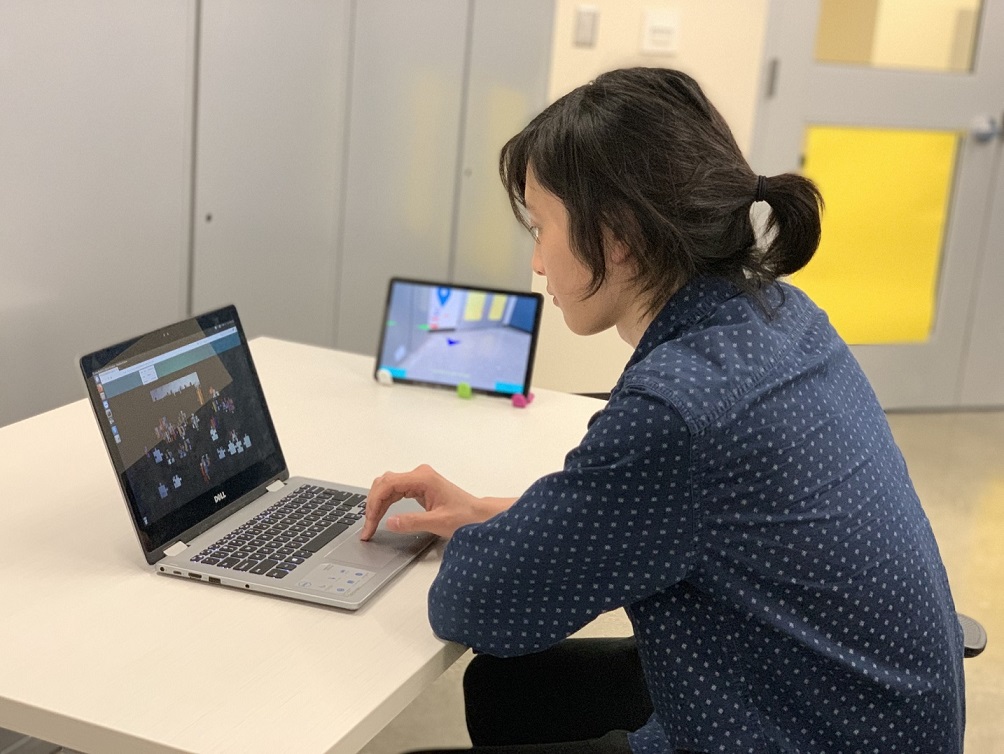}
    \label{fig:participant_solving_jigsaw}}
    \subfigure[Non-AR baseline]
    {\includegraphics[height=2.7cm]{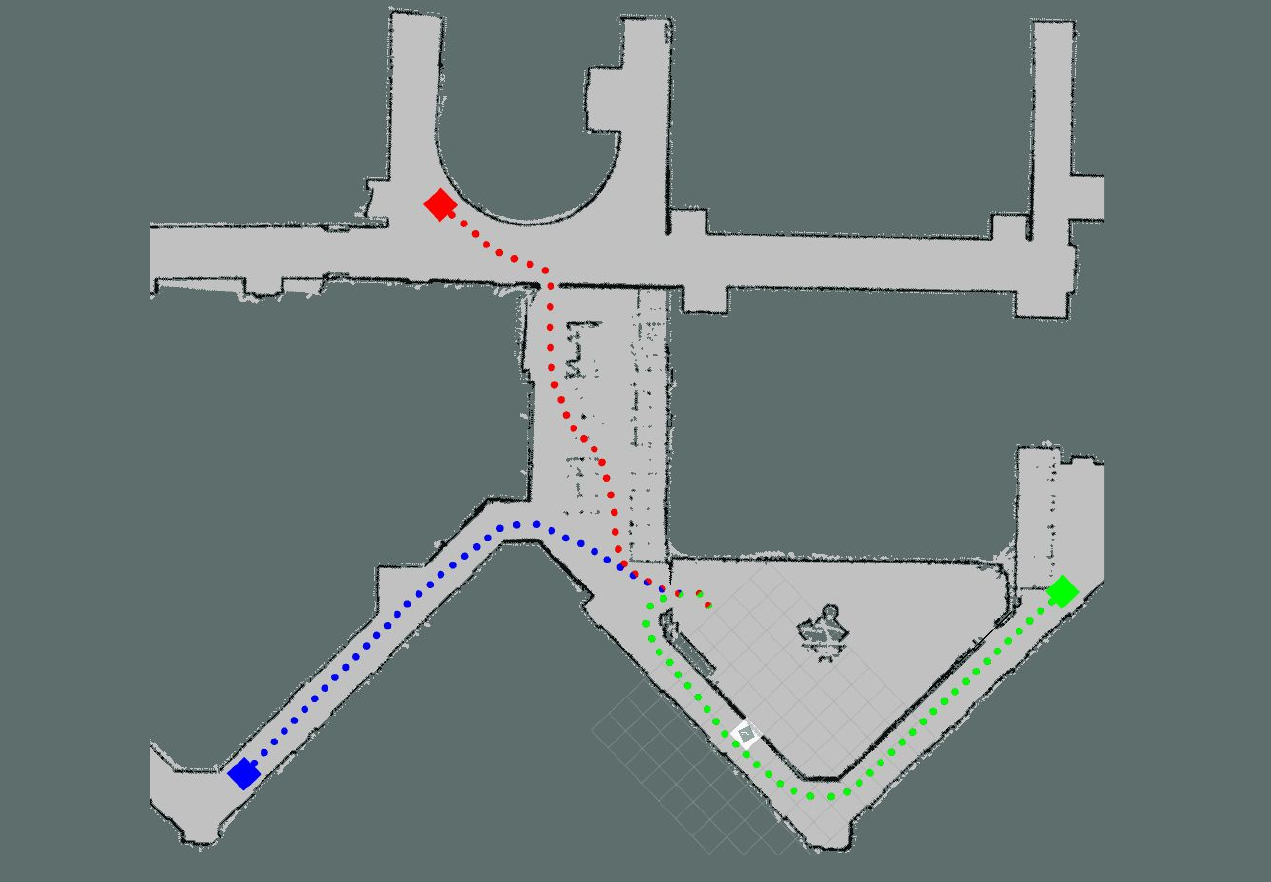}
    \label{fig:map}}
    \vspace{-.5em}
    \caption{(a)~A human participant playing the Jigsaw game alongside our AR interface running on a tablet, while helping the robots open the yellow door, where simulation environment shown in Fig.~\ref{fig:unity_table_chair} was constructed accordingly; and 
    (b)~The Rviz-based interface (non-AR baseline) showing the three robots' current locations and planned trajectories.
    }

    \label{fig:setup}
    \end{center}
    \vspace{-2.3em}
\end{figure}

We have conducted experiments with human participants collaborating with multiple robots in the real world. 
The tasks to the human-robot team are the same as those in simulation: robots work on delivery tasks, and the human plays Jigsaw puzzles while helping robots through door-opening actions. 
The puzzles at the same difficulty level were randomly generated in each trial to avoid the participants' learning behaviors. 
We selected the Jigsaw game due to its similarity to assembly tasks in the manufacturing industry. 
The \textbf{real-world setup} is shown in Fig.~\ref{fig:participant_solving_jigsaw}. 
We compared ARROCH to a standard \textbf{non-AR baseline}, where the human uses a Rviz-based interface to visualize the robots' current states and planned actions, as shown in Fig.~\ref{fig:map}.\footnote{http://wiki.ros.org/rviz}

\vspace{.5em}
\noindent{\bf{Participants:}}
Eleven participants of ages 20-30, all students from Binghamton University (BU), volunteered to participate in the experiment, including four females and seven males.
Each participant conducted two trials using both the Rviz-based baseline, and ARROCH -- randomly ordered. 
None of the participants had any robotics background. 
There was no training performed on ARROCH or the baseline prior to the trials. 
The experiments were approved by the BU Institutional Review Board (IRB).

\vspace{.5em}
\noindent{\bf{Task Completion Time:}}
Fig.~\ref{fig:scatter_plot} shows \emph{individual task completion times} produced by ARROCH and the non-AR baseline. 
The $x$ and $y$ axes correspond to the human's task completion time, and the total of the individual robots' task completion times. 
The two ellipses show the 2D standard deviations. 
Results support Hypothesis-I that states ARROCH improves the human-robot team's task completion efficiency. 
To analyze the statistical significance, we sum up the team agents' individual completion times (both human and robots), and found that ARROCH performed \textbf{significantly better} than the non-AR baseline, with $0.01<p$-value$<0.05$. 

We also looked into the \emph{team task completion time}, which is determined by the time of the slowest agent (human or robot). 
Fig.~\ref{fig:histogram_real} shows the histogram, where the observation is consistent with that of individual task completion time. 
For instance, in most trials of ARROCH, the teams used $\leq 950$ seconds, whereas in most trials of non-AR the teams could not complete the tasks within the same time frame. 
We observed that the participants, who did not have a robotics background, frequently experienced difficulties in understanding the visualization provided by the Rviz-based interface, including the map and robot trajectories. 
More importantly, ARROCH allows the participants to focus on their own tasks by indicating their unavailability, which is particularly useful in the early phase, whereas Rviz-based communication is unidirectional.

\vspace{.5em}
\noindent{\bf{Questionnaires:}}
At the end of each trial, participants were asked to fill out a survey form indicating their qualitative opinion over the following items. 
The response choices were: 1 (Strongly disagree), 2 (Somewhat disagree), 3 (Neutral), 4 (Somewhat agree), and 5 (Strongly agree).
The questions include: 
Q1, \emph{The tasks were easy to understand}; 
Q2, \emph{It was easy to keep track of robot status}; 
Q3, \emph{I could focus on my task with minimal distraction from robot}; 
Q4, \emph{The task was not mentally demanding (e.g., remembering, deciding, and thinking)}; and 
Q5, \emph{I enjoyed working with the robot and would like to use such a system in the future.}
Among the questions, Q1 is a verification question to evaluate if the participants understood the tasks, and is not directly relevant to the evaluation of our hypotheses.

Fig.~\ref{fig:bar_graph} shows the average scores from the questionnaires.
Results show that ARROCH produced higher scores on Questions Q2-Q5, where we observed \textbf{significant improvements} in scores in Q2, Q3, and Q5 with the $p$-values $<0.001$.
The significant improvements support Hypothesis-II on user experience: ARROCH helps keep track of the robot status, is less distracting, and is more user-friendly. 
The improvement in Q4 was not significant, and one possible reason is that making quantitative comparisons over the ``mentally demanding'' level can be difficult for the participants.


\section{Conclusions}
Leveraging augmented reality (AR) technologies, we introduce \emph{AR for robots collaborating with a human} (ARROCH), a novel algorithm and system that enables bidirectional, multi-turn, beyond-proximity communication to facilitate collaborative behaviors within human-multi-robot teams.
ARROCH enables the human to visualize the robots' current states and their intentions (planned actions), while supporting feedback to the robots. 
The human feedback is then taken by the robots toward human-robot collaboration.
Experiments with human participants showed that ARROCH performed better than a traditional non-AR approach, while simulation experiments highlighted the importance of ARROCH's bidirectional communication mechanism. 



\section*{Acknowledgment}
A portion of work has taken place in the Autonomous Intelligent Robotics (AIR) Group at SUNY Binghamton. AIR research is supported in part by grants from the National Science Foundation (NRI-1925044), Ford Motor Company (URP Awards 2019 and 2020), OPPO (Faculty Research Award 2020), and SUNY Research Foundation.

\bibliographystyle{IEEEtran}
\bibliography{ref}

\end{document}